# Surrogate Quality Score: A Novel Composite Performance Metric for Model Selection

Maaz Ahmad and Iftekhar A Karimi[1]
*Department of Chemical & Biomolecular Engineering*
*National University of Singapore*
*4 Engineering Drive 4, Singapore 117585*

**Abstract**

The growing reliance on high-fidelity, state-of-the-art machine learning based models for engineering decision-making necessitates robust and systematic approaches for model selection. While high predictive accuracy at untrained points is a necessary trait in the final model, other attributes such as model complexity, forecastability, explainability, scalability, and physics-awareness may be equally important or relevant criteria depending on application context and user requirements. In this work, we propose a novel composite performance metric, Surrogate Quality Score (*SQS*) for model selection based on predictive accuracy and model complexity simultaneously. *SQS* heuristically integrates these two attributes through a weighted product of normalized measures, yielding a unified score to reflect overall model quality. A user-specified parameter in *SQS* enables explicit control over the accuracy-complexity trade-off, thereby incorporating human judgement and preference in the model selection process. We extensively discuss and justify the formulation of *SQS* and demonstrate its utility through an illustrative example. By combining predictive accuracy and model complexity, *SQS* offers a systematic and objective approach for model selection based on both attributes.

[1] Corresponding author: Email cheiak@nus.edu.sg

## 1. Introduction

Rapid advancements in data handling and storage, development of high-speed computer processors, improved sensor technologies aided by Internet-of-Things, and the (re)-emergence of artificial intelligence (AI) have fostered significant interest in the domain of data-driven machine learning (ML) for scientific discoveries and engineering applications (Leonard et al., 2021; Mann et al., 2025; Venkatasubramanian, 2019). ML models establish correlations between the input and output / response variables of a physical or in-silico system by learning its underlying patterns from input-output data. The data for this model training process needs to be sufficient in both quantity and quality to effectively capture the system characteristics (relationships, dynamics, nonlinearities). However, generating large volumes of high-quality data is often impractical in real-world settings. Depending on the nature of the data-generating system, the primary limitations arise from labor-intensive experimentations, compute-intensive simulations, or infeasible, costly, and/or risky plant operations. Several design of experiments (DoE) or sampling methodologies (Ahmad et al., 2026; Crombecq et al., 2009; Eason and Cremaschi, 2014; Farhang-Mehr and Azarm, 2005; Garud et al., 2017; Wang and Ierapetritou, 2017) have addressed these challenges to extract maximal system information from minimal sampling.

The other critical aspect of ML modeling is model selection. This involves selecting a modeling technique (learning algorithm) and an analytical form for the ML model. Model selection has been a topic of much research even before the advent of AI. Classical model selection (Chamberlin, 1965; Forster and Sober, 1994; Forster, 2000) is rooted in the premise that no model is perfect. All models are imperfect realizations or mere approximations of reality (ground truth). The "no free lunch" theorem in statistics (Wolpert, 1996) asserts that no single model dominates all others across all possible data sets. In other words, considering and assessing one or more working hypotheses, models, or modeling techniques is crucial for most real systems. Given the vast variety of ML techniques and several and often competing/conflicting traits of a desirable model, the key question is: What is the "best" model?

In practice, most non-ML users select models solely based on predictive accuracy or the ability of a model to predict system response accurately beyond seen or train points. It is normally assessed at some unseen or test points. Several recent works (Ahmad and Karimi, 2021; Chakraborty et al., 2021; Cozad et al., 2014; Garud et al., 2018; Williams and Cremaschi, 2021) have developed automated frameworks for selecting models based on predictive accuracy. However, several other attributes such as model complexity, interpretability / explainability, scalability, forecasting ability, and physics-compatibility are quite relevant and desirable in modeling real-world systems. Fig. 1 highlights their impacts on and correlations with predictive accuracy schematically. A savvy modeler would normally prioritize, assess, and optimally trade them off in the "best" model. However, this task is system-dependent, highly subjective, and certainly not straightforward due to the multiple diverse, disparate, and conflicting attributes. It invariably demands foundational ML knowledge, domain understanding, and human judgment.

The case of model complexity is quite relevant and interesting. While good predictive accuracy requires a sufficiently complex model, greater model complexity does not necessarily mean higher predictive accuracy. An overly simplistic and inflexible model cannot capture the behavior of a large, complex, nonlinear system adequately, irrespective of the quality and quantity of training data. Such a model underfits the training data and performs poorly at both the train and test points. Conversely, an overly complex model may learn spurious correlations (e.g. response noise) and non-existent interactions and may overfit the training data. Such a model fits the train points very well but does poorly at test points. This reality is captured by the well-established bias-variance tradeoff (Hastie et al., 2009) that can be measured in terms of $ESS$ or the expected value of squared error between true responses and model predictions, where the expectation is across many train sets. $ESS$ has three components as follows.

$$ESS = Bias^2 + Variance + \sigma_\epsilon^2 \quad (1)$$

$Bias$ is the systematic error that an insufficiently complex model form fails to capture from the training data. $Variance$ is the error that arises due to the variations in training data sets. $\sigma_\epsilon^2$ is the inherent system noise that cannot be reduced by varying model complexity. Underfitting models suffer from high $Bias$ but low $Variance$, while overfitting models exhibit low $Bias$ but high $Variance$. Both result in high $ESS$. In general, $Bias$ decreases and $Variance$ increases with model complexity, resulting in a characteristic U-shaped $ESS$ curve (Fig. 2). The minimum of this curve (marked by blue star in Fig. 2) gives us the optimal model complexity that balances the bias-variance tradeoff. Since predictive accuracy tries to estimate $ESS$, it does address model complexity, but partially and implicitly. An explicit consideration of predictive accuracy AND unaddressed aspects of model complexity remain largely ignored in model selection.

Selecting a model of appropriate complexity is critical in many real-world industrial applications such as online process control, continuous monitoring, and real-time model adaptation, and optimization. The idea of preferring a simple model to a more complex one is not new. The principle of parsimony (Gauch, 2012) recommends preferring the simplest model from equally accurate models. The one standard error rule (Hastie et al., 2009) furthers the principle of parsimony for *similarly accurate* models. It recommends selecting the simplest model from those whose expected accuracy lies within one standard deviation of the most accurate model. However, domain experts may (re)-define subjective notions such as "similar accuracy" and "sufficient accuracy" based on their system understanding and modeling requirements. They may exercise greater discretion in the model selection process by pre-specifying requirements such as thresholds for minimum accuracy and/or maximum complexity. Their system knowledge and modeling needs may even prompt them to sacrifice some predictive accuracy in exchange for a simpler model, even if the candidate models are NOT similar or sufficiently accurate. The main challenge is that despite being partially correlated, model complexity and predictive accuracy are two disparate measures that cannot be placed on the same scale. Balancing them is not trivial, so a way to integrate them together would be highly desirable in selecting the best model based on both attributes collectively.

One established approach for selecting the best model in the presence of multiple, potentially conflicting attributes is to employ multi-attribute decision making (MADM), multi-criteria decision making (MCDM), or multi-criteria decision analysis (MCDA) (Triantaphyllou, 2000). They help identify (Fig. 3) the best solution from several competing alternatives (candidate solutions) based on multiple attributes or evaluation criteria through a common sequence of steps such as identifying relevant attributes and alternatives, normalizing attributes to a common range, assigning weights to attributes, deriving a consolidated and weighted score, etc. However, their key feature is the incorporation of human judgement at various steps. Several recent works (Correia and Lecue, 2019; Ghosh and Bequette, 2020; Savage and Del Rio-Chanona, 2024) have also recognized the unavoidability of human participation in AI-powered workflows for meaningful, reliable, and informed decision-making. This collaborative paradigm, termed as *human-in-the-loop AI* (HITL-AI) (Mosqueira-Rey et al., 2023) has gained prominence for its ability to combine AI-based decisions with human reasoning, intuition, and system understanding.

In this work, we focus on balancing predictive accuracy versus model complexity as the two key attributes for model selection, and develop an objective, rational, and systematic approach for the same. While our approach adopts MADM/MCDM in spirit, it minimizes the user specifications for various open-ended choices typical of generic MADM/MCDM methods. It relies completely on a novel, compact, hybrid performance metric, called Surrogate Quality Score (*SQS*), which integrates predictive accuracy and model complexity into a single quantitative score. *SQS* embeds a tunable parameter that can be tweaked by human experts to adjust the accuracy-complexity tradeoff. By including human experts into the model selection process, *SQS* selects the best model that is both data-driven and informed by human judgement, intuition, system understanding, and modeling requirements.

It is important to mention that *SQS* is fundamentally different from existing performance metrics that consider the impact of a measure of model complexity on predictive accuracy, such as Mallows Cp (Mallows, 1973), Akaike Information Criterion (AIC) (Akaike, 1974), and Bayesian Information Criterion (BIC) (Schwarz, 1978). These metrics combine goodness-of-fit (accuracy at train points) with model complexity to *estimate* predictive accuracy (at test points) of a model relative to other candidate models. We will discuss this later in more detail. In contrast, *SQS* is a heuristic metric that is derived by careful consideration of several salient features. A crude version of *SQS* was incorporated in our automated framework for model selection, LEAPS2v2 (Ahmad and Karimi, 2021). However, it lacked user control or human input. In this work, we significantly enhance that initial version based on extensive analysis and feedback from users. The updated *SQS* (henceforth called *SQS*) uses a modified formulation to better compare candidate models across diverse modeling techniques. *SQS* can specifically guide domain experts in real-world settings (industrial manufacturing, scientific research, etc.) to select models that are both sufficiently accurate and practically simple to best meet their desired modeling objectives. To our best knowledge, no literature metric has explicitly combined predictive accuracy and model complexity in this manner.

The remainder of this paper is structured as follows. The next section details the primary objective and criteria for model selection. Section 3 briefly reviews various methods and metrics to compute model complexity. Section 4 focuses on predictive accuracy along with a discussion of complexity-based metrics (Mallows Cp, Stein's Unbiased Risk Estimate, and AIC) and their limitations. Section 5 presents our proposed metric *SQS*, highlighting its salient features and outlining a systematic approach for applying it to model selection. Finally, we conclude with a discussion on possible improvements and future directions for ML-based model selection.

## 2. Our Objective

Let $\boldsymbol{x} = [x_1, x_2, \dots, x_N] \subset \mathbb{R}^N$ denote an input or sample point constituting $N$ input variables and $f(\boldsymbol{x} \in D): \mathbb{R}^N \to \mathbb{R}$ denote the system output or response based on the true functional relationship $f$ between the inputs and output. The observed response $y(\boldsymbol{x})$ can either equal $f(\boldsymbol{x})$ or can have some white noise $\epsilon$ added to $f(\boldsymbol{x})$. Let $S(\boldsymbol{x})$ denote an ML-based surrogate model trained on a train set $\mathcal{T}: [\boldsymbol{x}^{(k)}, y^{(k)}], k = 1, 2, \dots, K$ comprising $K$ sample points to approximate $f(\boldsymbol{x})$. Our overall objective is to propose a heuristic quantitative metric for model selection, which integrates the attributes of predictive accuracy and model complexity along with the following desirable traits.

**T1.** It favors a model with high predictive accuracy.

**T2.** It favors a model with low complexity.

**T3.** It tries to follow the principle of parsimony.

**T4.** It can be used for any modeling technique or model form.

**T5.** It provides a finite and bounded measure of overall model quality.

**T6.** It respects a user-specified notion of 'sufficient accuracy'.

**T7.** It allows human experts to tweak accuracy-complexity tradeoffs.

The next two sections discuss the existing approaches to quantify model complexity and predictive accuracy and lay the foundation for our *SQS*.

## 3. Measuring Model Complexity

A straightforward, but naïve, measure of model complexity is the number of fitted parameters in a model. While easy to calculate, this measure can be misleading for two key reasons. First, it ignores inter-parameter dependence and interactions. In some modeling techniques (Radial

Basis Functions, Artificial Neural Networks, etc.), some parameters directly influence others. Thus, not all parameters may be fully independent or contribute equally to a model's flexibility. Second, it does not consider the constraints imposed on the parameters during model training. Techniques like regularization or sparsity penalties (Girosi et al., 1995), early stopping (Yao et al., 2007), and parameter clipping (Elsayed et al., 2024) restrict parameters from achieving their unconstrained optimal values. These constraints effectively reduce the flexibility of a trained model. Thus, the number of fitted model parameters may overestimate model complexity.

A better measure of model complexity is its degrees of freedom ($dof$) as proposed by (Efron, 1986), which estimates the *effective* number of parameters in a model. However, it assumes that the system responses have noise: $y = f(\boldsymbol{x}) + \epsilon$ with $E(\epsilon) = 0$ and $Var(\epsilon) = \sigma_\epsilon^2$. Note that $Var(y) = \sigma_\epsilon^2$. Consider a data set $\mathcal{T}{:}\left[\boldsymbol{x}^{(k)}, y^{(k)}\right], k = 1, 2, \dots, K$ and a model $S(\boldsymbol{x})$ trained on $\mathcal{T}$. The degrees of freedom $dof(S)$ of $S$ measures how flexibly it can adapt to the variations in observed responses as follows,

$$dof(S) = \frac{1}{\sigma_\epsilon^2} \sum_{k=1}^{K} Cov\left(y^{(k)}, \hat{f}^{(k)}\right) \tag{2}$$

where, $Cov\left(y^{(k)}, \hat{f}^{(k)}\right)$ denotes covariance and $dof$ provides a reliable and intuitive measure of the effective number of parameters in a model. For instance, a simple average model $S(\boldsymbol{x}) = \frac{1}{K}\sum_{k=1}^{K} y^{(k)}$ has $dof(S) = 1$. Conversely, a model that perfectly interpolates each train point as $S(\boldsymbol{x}) = y$ has $dof(S) = K$. This is in line with our logical understanding that an interpolative model has the highest flexibility or complexity. For a linear model with $N + 1$ parameters ($N$ inputs and 1 constant), $dof = N + 1$. However, for a regularized linear model (where parameters are constrained), $dof < N + 1$ and depends on the strength of regularization. This reduction in $dof$ is again intuitive as the constraints on model parameters reduce its flexibility.

A major difficulty with $dof$ is that fixing it analytically is not possible for all models except those whose parameters can be expressed as functions of observed responses $y^{(k)}$. In most cases, $dof$ must be estimated empirically via Eq. 2 as follows. First, estimate $\sigma_\epsilon^2$ by either evaluating repeat responses or using system knowledge. Then, obtain several $(B \gg 1)$ evaluations of system responses $y^{(k)}$, $k = 1, 2, \ldots, K$ to prepare $B$ distinct data sets, each with the same set of inputs but different responses. Train a unique model on each data set and obtain its predictions at the $K$ train points. Compute the covariance between the true response and model prediction at each sample point across $B$ evaluations to get an estimate of $dof$ using Eq. 2. Strictly speaking, $dof$ is undefined for deterministic (simulation) responses, and one would have to artificially impose a white noise and then estimate $dof$ using the above procedure.

Another measure of complexity, namely the Vapnik-Chervonenkis Dimension or $vcd$ was proposed by (Vapnik and Chervonenkis, 1971), using the concept of 'shattering'. Given sample points with random binary labels (0-1), a model is said to 'shatter' all points, if it can successfully predict (perfectly classify) any labeling scheme on the points. $vcd$ of a model form is the largest number of sample points that can be shattered by a model derived from that form. While the VC dimension is natural for classification models (those that predict labels), it needs a prediction-to-label transformation for regression models (those that predict real-valued responses). However, unlike $dof$, $vcd$ is defined for a model form (class of models), rather than a trained instance of a model. Computing $vcd$ for complex model forms with several (tunable and hyper) parameters (such as deep neural networks, adaptive splines, etc.) is challenging due to the numerous combinatorial assessments associated with several possible models. See (Hastie et al., 2009; Vapnik and Chervonenkis, 1971) for a deeper discussion of $vcd$.

## 4. Computing Predictive Accuracy

Recall that predictive accuracy measures the ability of a model to generalize beyond train points, or specifically at test points. Literature offers at least five classes (A, B, C, D, and E) of methods to quantify predictive accuracy.

Class A uses two distinct data sets: a train set $\mathcal{T}$ for constructing the model $[S(\boldsymbol{x})]$, and a test set $\mathcal{T}'$: $\left[\boldsymbol{x}^{(k)}, y^{(k)}\right], k = 1, 2, \ldots, K'$ for computing its predictive accuracy. The predictive accuracy is defined via one or more error metrics listed below.

$$Mean\ Absolute\ Error\ (MAE) = \frac{1}{K'}\sum_{k=1}^{K'}\left|y^{(k)} - \hat{f}^{(k)}\right| \tag{3}$$

$$Mean\ Squared\ Error\ (MSE) = \frac{1}{K'}\sum_{k=1}^{K'}\left(y^{(k)} - \hat{f}^{(k)}\right)^2 \tag{4}$$

$$Root\ MSE\ (RMSE) = \sqrt{\frac{1}{K'}\sum_{k=1}^{K'}\left(y^{(k)} - \hat{f}^{(k)}\right)^2} \tag{5}$$

$$Predictive\ R^2\ (PredR^2) = 1 - \frac{K' \times MSE}{\sum_{k=1}^{K'}(y^{(k)} - \bar{y})^2} \tag{6a}$$

$$\bar{y} = \frac{1}{K'}\sum_{k=1}^{K'} y^{(k)} \tag{6b}$$

Unlike class A, classes B-E use only one data set $\mathcal{T}''$with $K''$ points. Class B (Holdout Validation or HV) splits $\mathcal{T}''$ randomly into train and test subsets, with the former typically being larger than the latter. Then, the predictive accuracy is computed using Eq. 3-6. The process can be repeated to obtain an average, more representative, estimate of predictive accuracy.

Class C (Cross-Validation or CV) partitions $\mathcal{T}''$ into $F \geq 2$ exclusive and exhaustive subsets and constructs $F$ distinct models by considering each subset one by one as a hold-out subset. At each iteration, the hold-out subset serves as the test set, and the other subsets collectively as the train set. The predictive accuracy is simply the average accuracy of $F$ models. This version (Stone, 1974) of CV (called F-fold CV) is the most common. A special case of this is Leave-one-out CV or $K''$-fold CV. In practice, an F-fold CV with $5 \leq F \leq 10$ offers a good compromise between reliability and computational efficiency.

Class D (Bootstrapping or BS) (Efron, 1979) generates $B \geq 2$ bootstrapped subsets by sampling some points from $\mathcal{T}''$ with replacement. At each iteration, the bootstrapped subset serves as the train set, and the unsampled points (out-of-bag subset) serves as the test set. Predictive accuracy is averaged over the B iterations/models.

Class E (Penalized Goodness-of-Fit) trains uses the entire $\mathcal{T}''$ as the train set, and estimates predictive accuracy based on the model's goodness-of-fit (*gof*) using statistical arguments. As discussed earlier, prediction errors at test points are systematically higher than at train points. Past works estimated this upward bias via several metrics and added it to $gof$ to get predictive accuracy. (Mallows, 1973) proposed the following predictive accuracy for linear models, where the first term is $gof = MSE$ on the train set and the second is the upward bias.

$$Cp\ Statitsic\ (for\ linear\ models) = \frac{1}{K}\sum_{k=1}^{K}\left(y^{(k)} - \hat{f}^{(k)}\right)^2 + \frac{2(N+1)}{K}\hat{\sigma}_\epsilon^2 \quad (7)$$

In the above, $\hat{\sigma}_\epsilon^2$ denotes an estimate of the noise variance. (Efron, 2004) generalized the Cp-statistic to arbitrary model forms (linear or nonlinear) by expressing the bias in terms of $dof$ as follows (Eq. 2).

$$Generalized\ Cp\ Statitsic\ (for\ all\ models) = \frac{1}{K}\sum_{k=1}^{K}\left(y^{(k)} - \hat{f}^{(i)}\right)^2 + \frac{2dof}{K}\sigma_\epsilon^2 \quad (8)$$

(Stein, 1981) assumed a Gaussian system noise, $\epsilon \sim N(0, \sigma_\epsilon^2)$, and proposed SURE (Stein's Unbiased Risk Estimate), which is valid for both linear and nonlinear models.

$$SURE = \frac{1}{K}\sum_{k=1}^{K}\left(y^{(k)} - \hat{f}^{(k)}\right)^2 + \frac{2\sigma_\epsilon^2}{K}\sum_{k=1}^{K}\frac{\partial \hat{f}^{(k)}}{\partial y^{(k)}} \quad (9)$$

Note that replacing $dof$ by $\sum_{k=1}^{K}\frac{\partial \hat{f}^{(k)}}{\partial y^{(k)}}$ in eq. 8 gives us eq. 9. (Akaike, 1974) proposed AIC (Akaike Information Criterion) for models whose parameters are estimated by the maximum likelihood estimation (MLE) technique. MLE (Myung, 2003) offers a probabilistic framework for estimating model parameters by assuming a probability distribution function for the train error. The *gof* of a model (with parameters $\widehat{\boldsymbol{\theta}}$) trained by MLE is quantified by the $Likelihood$ function as follows, which computes the joint probability of observing the training responses from a trained model.

$$Likelihood = \prod_{k=1}^{K} \Pr\left(y^{(k)} \middle| x^{(k)}; \widehat{\boldsymbol{\theta}}\right) \tag{10a}$$

(Akaike, 1974) showed that the probability of observing the true responses at a test point can be obtained as follows.

$$E\left[\log\left(\Pr y|x; \widehat{\boldsymbol{\theta}}\right)\right] = \frac{1}{K}\left[\log(Likelihood) - dof\right] \tag{10a}$$

Multiplying the above by $-2K$ yields AIC that serves as a relative measure of predictive accuracy among competing models.

$$AIC = 2dof - 2\log(Likelihood) \tag{10b}$$

For a model with Gaussian-distributed train errors, AIC simplifies to,

$$AIC = K\ln(\hat{\sigma}_\epsilon^2) + K(\ln(2\pi) + 1) + 2dof \tag{10c}$$

where, $\hat{\sigma}_\epsilon^2$ is estimated as $MSE$ on the train set.

## 5. Surrogate Quality Score (*SQS*)

Let $\boldsymbol{S}: \{S_1(\boldsymbol{x}), S_2(\boldsymbol{x}), \dots, S_M(\boldsymbol{x})\}$ denote a pool of $M$ surrogate models. Let us use $MSE_m$ (Mean Squared Error) as a measure of predictive accuracy and $dof_m$ as a measure of model complexity for model $m$. Let $s_y^2$ denote the sample variance in responses at the train points as follows.

$$s_y^2 = \frac{1}{K-1}\sum_{k=1}^{K}\left[y^{(k)} - \bar{y}\right]^2 \quad with \quad \bar{y} = \frac{1}{K}\sum_{k=1}^{K} y^{(k)} \tag{11}$$

Define $MSE^*$ as an optional user-specified maximum $MSE$ for a model to be considered sufficiently accurate, and $\boldsymbol{S}^* \subseteq \boldsymbol{S}$ as the subset of sufficiently accurate models: $\boldsymbol{S}^* = \{S_m(\boldsymbol{x}) \in \boldsymbol{S} \mid MSE_m \leq MSE^*\}$.

We combine $MSE$ (predictive accuracy) and $dof$ (model complexity) into a composite score called $SQS$ for assessing overall model quality as follows.

$$SQS_m = \frac{(CF_m)^\gamma}{s_y^2}\left[\widehat{MSE} + \max\left(0, MSE_m - \widehat{MSE}\right)\right] \qquad 0 < SQS_m \qquad (12a)$$

$$CF_m = \left[\frac{\ln(1 + dof_m)}{\ln\left(1 + \max\limits_{m'} dof_{m'}\right)}\right] \qquad (12b)$$

$$\widehat{MSE} = \operatorname*{median}_{SQS_m \in \boldsymbol{S}^*} MSE_m \qquad (12c)$$

We call $CF_m$ $(0 \leq CF_m \leq 1)$ the complexity factor. It normalizes the contribution of $dof$ to [0, 1]. The exponential factor $\gamma \geq 0$ is a user-tunable parameter to allow a human expert to tweak the accuracy-complexity tradeoff. Note that *SQS* is a comparative or relative metric versus an absolute measure like $MSE_m$ or $dof_m$. The lower the $SQS_m$, the better the model.

Given a pool $\boldsymbol{S}$ of models, selecting the best model based on *SQS* would proceed as follows. If any two models in $\boldsymbol{S}$ have the same $dof$, then discard the model with the higher *MSE*. Repeat the procedure until no two models have the same $dof$. This gives us a revised pool $\boldsymbol{S}$. Now, compute *SQS* for all models in the revised $\boldsymbol{S}$, and select the one with the lowest $SQS$.

*SQS* incorporates three key steps of a typical MADM/MCDM exercise, namely attribute specification and normalization, weight assignment, and consolidation scheme. It normalizes $MSE_m$ using $s_y^2$ and $dof_m$ via $CF_m$. Through this, it brings them both to a common scale. Instead of assigning distinct weights to each attribute, *SQS* uses a single weight / parameter $\gamma$ to capture the relative importance of model complexity with respect to predictive accuracy. Finally, its consolidation scheme involves multiplying both attributes into a single score, which resembles the weighted product model (WPM) commonly employed in MADM/MCDM (Triantaphyllou, 2000).

$SQS$ embodies the desirable traits (T1 – T7) listed in Section 2. It assigns lower scores to models with higher predictive accuracy (T1) and lower model complexity (T2). It respects the principle of parsimony (T3) by preferring the simplest from among equally accurate models. It is generic (T4) and not limited to specific model forms or parameter estimation methodologies. It is a dimensionless scaled metric (T5). It incorporates a user-defined threshold for acceptable accuracy, which allows us to favor simpler ones among models that are sufficiently accurate. (T6). Finally, it allows users to adjust the accuracy-complexity tradeoff via $\gamma$ (T7). $\gamma = 0$ reduces $SQS$ to a purely accuracy-based metric.

### 5.1 Justification for $SQS$ Form

$SQS$ is essentially a heuristic metric that integrates two key model aspects (predictive accuracy and model complexity) in a product form. Multiple options exist for the two aspects and their integration. Hence, the rationale or thought processes underlying our construction needs some explanation.

***<u>MSE for predictive accuracy</u>***

We discussed several metrics for predictive accuracy in Section 4. In principle, any of them, except AIC, could be used in place of $MSE$. We used $MSE$ primarily due to its interpretability and familiarity. It can be computed easily from $RMSE$ (Eq. 5) and predictive-$R^2$ (Eq. 6), while the Generalized Cp-statistic (Eq. 8) and SURE (Eq. 9) provide direct estimates of $MSE$. Moreover, $MSE$ relates well with the user-specified $MSE^*$. Incorporating AIC in $SQS$ poses two fundamental challenges. First, strictly speaking, AIC is applicable only to models whose parameters are estimated using the MLE technique. This would restrict the applicability of $SQS$ to a specific class of models and contradict our objective (T4) of maintaining generality across modeling techniques and forms. Second, AIC provides a *relative* measure of predictive accuracy based on the Likelihood function, which interprets differently from the *absolute* error-based metrics (such as MAE, MSE, RMSE, $R^2$, Cp-statistic, SURE) discussed in Section 4. Thus, a separate normalization scheme for AIC alone would be required within $SQS$.

***$dof$ for model complexity***

Among the three options ($dof$, $vcd$, number of model parameters) discussed in Section 3 for model complexity, $dof$ offers a sound, theoretical, and realistic estimate. As stated before, $vcd$ is applicable to model classes or forms rather than individual trained models, and the number of free model parameters can overestimate model complexity.

***$CF$ for a more intuitive and transformed measure of complexity***

By employing a logarithmic transformation of $dof$, $CF$ addresses two key aspects. First, the impact of a given increment in $dof$ depends on the level of $dof$. For instance, the effect of increasing $dof$ from 5 to 10 is much greater than increasing it from 100 to 105 or 1000 to 1005. A logarithmic transformation of $dof$ captures this intuition. Second, a linear scale for $dof$ may exaggerate the preference for model simplicity and over-penalize complex yet accurate models. By compressing the range of $CF$, the logarithmic transformation helps achieve a better balance between predictive accuracy and model complexity. Fig. 4 illustrates these two aspects on a $\ln(dof)$ vs $dof$ plot.

***$\ln(1 + dof)$ versus $\ln(dof)$ in complexity factor***

$CF$ normalizes $\ln(1 + dof)$ rather than $\ln(dof)$ to avoid assigning $SQS = 0$ to the simplest model with $dof = 1$. This ensures that even the simplest model has a nonzero complexity factor and a non-zero aggregate score for $SQS$.

***Separating noise from accuracy***

$SQS$ recognizes the undesirability of a model that tries to fit noise. Eq. 12a sets a user-defined threshold for our practical ability to explain real physical data. Hence, it treats all models as equal in terms of accuracy as long as they meet this threshold. This is quite important for real-life experimental data. A common representative accuracy measure ($\widehat{MSE}$) for all models in $\boldsymbol{S}^*$ ensures that $SQS$ ranks them solely based on model complexity, while being comparable with models outside $\boldsymbol{S}^*$.

***<u>Product form aggregates normalized measures of accuracy and complexity</u>***

By aggregating predictive accuracy and model complexity in a product form, $SQS$ avoids forcing them on one single scale. Besides simplicity, the product form ensures that a model with $MSE_m = 0$ (true model) always has the best score ($SQS_m = 0$).

### 5.2 Choosing $\gamma$

Since $\gamma$ is user-specified, some plausible suggestions for its selection are deserve discussion.

#### 5.2.1 Quantify the Accuracy-Complexity Trade-off

One may quantify a user's desired accuracy-complexity trade-off by asking: *If a model's predictive accuracy (MSE) reduces by a known factor (say $\alpha > 1$), then what factor increase (say $\beta > 1$) in complexity ($\ln[dof]$) can they tolerate, while still considering it the best model?* The user's response can then be expressed mathematically in $\gamma$ as follows.

Let $MSE_{m^*}$ and $CF_{m^*}$ denote the accuracy and complexity factors for the best model. Let another model have $MSE = MSE_{m^*}/\alpha$ and $CF = \beta CF_{m^*}$. If both models are to have the same $SQS$, then the following must hold:

$$\frac{MSE_{m^*}}{\alpha}[\beta CF_{m^*}]^{\gamma} = MSE_{m^*}[CF_{m^*}]^{\gamma} \qquad or \qquad \gamma = \frac{\ln(\alpha)}{\ln(\beta)} \tag{13}$$

A user may have three possible perspectives on the accuracy-complexity trade-off, which can be translated into a specific value for $\gamma$ as follows:

**User favors predictive accuracy strongly.** If we assume $\alpha = 2$, then we can take $\beta > 2$. For instance, $\beta = 10$ would mean that the user would prefer a model that is 10 times more complex but doubles accuracy. This yields $\gamma \cong 0.3$.

**User favors model simplicity strongly.** In this case, we should take $\beta < 2$ for $\alpha = 2$. For instance, $\beta = 1.3$ would mean that the user would prefer a model that doubles accuracy, only if it is not more than 30% complex. This yields $\gamma \cong 2.64$.

**User favors both accuracy and complexity.** In this case, we can take $\alpha = \beta = 2$ to get $\gamma = 1$. However, $\alpha$ and $\beta$ need not be the same if users seek a balanced trade-off. For instance, $\alpha = 2$ and $\beta = 3$ yields $\gamma \cong 0.63$. This invokes comparison with the well-known six-tenths rule (Tribe and Alpine, 1986) related to the economies of scale. This rule relates the capital cost ($C$) of an equipment / plant to its operational capacity ($V$) as $C \propto V^{0.6}$. The sub-linear increase in $C$ with $V$ reflects the diminishing marginal cost with increasing capacity. By analogy, we can view model complexity as capacity or size and *SQS* as cost. Then, although a heuristic, $\gamma = 0.6$ may serve as a rational and pragmatic choice.

#### 5.2.2 Use Established MADM/MCDM Weight Assignment Methods

A more structured and objective approach for fixing $\gamma$ is to adopt weight assignment techniques commonly employed in MADM/MCDM analyses. These include CRITIC (CRiteria Importance Through Intercriteria Correlation) (Diakoulaki et al., 1995), MEREC (MEthod based on Removal Effects of Criteria) (Mehdi et al., 2021), entropy method (Mon et al., 1994), Gini coefficient (Li and Chi, 2009), etc. Their underlying idea is to examine the dispersion or variability of attribute values across candidate solutions and correlation among the attributes. Attributes with higher variability possess greater discriminatory power and are assigned larger weights. The different weight assignment methods differ in the metrics that quantify attribute variability. Interested readers can refer to the review articles by (Wang et al., 2024) and (Triantaphyllou, 2000) for a detailed description of these methods.

Two important points deserve mention. Firstly, all weight assignment methods determine a distinct weight for each attribute. For *SQS*, this would give separate weights, $\gamma_{acc}$ and $\gamma_{comp}$ for predictive accuracy and model complexity, respectively. This is equivalent to using $\gamma =$ $\gamma_{comp}/\gamma_{acc}$ in $SQS$ to obtain the same ranking scheme as derived by the weighted product model of MADM/MCDM given by $MSE^{\gamma_{acc}} \times CF^{\gamma_{comp}}$. Secondly, since the attribute weights depend on the distribution of attribute values within the available candidate solutions, modifying the pool of candidate models may alter $\gamma$.

**6. Illustrative Example**

We now illustrate the role of $\gamma$ in $SQS$ for model selection by approximating the two-dimensional Ackley function (Eq. 14) over $D \in [-32.768, 32.768]^2$.

$$f(x_1, x_2) = -20 \exp\left[-0.2\sqrt{\frac{x_1^2 + x_2^2}{2}}\right] - \exp\left[\frac{cos(2\pi x_1) + cos(2\pi x_2)}{2}\right] + 20 + \exp(1) \quad (14)$$

Its response surface in Fig. 5 shows a distinct symmetrical valley at the domain center.

Consider four surrogate model forms in $\boldsymbol{S}$ to approximate $f(\boldsymbol{x})$: a second-degree Polynomial Response Surface Model (PRSM) form ($S_{PRSM}$) (Eq. 15), a Radial Basis Function (RBF) form ($S_{RBF}$) with multi-quadratic basis function (Eq. 16), an Artificial Neural Network (ANN) form ($S_{ANN}$) with 1 hidden layer, 8 hidden nodes, and hyperbolic tangent (*tanh*) activation function (Eq. 17ab), and a Random Forest (RF) form ($S_{RF}$) built using $T = 100$ decision trees (Eq. 18).

$$S_{PRSM}(\boldsymbol{x}) = \beta_0 + \sum_{n=1}^{N} \beta_n x_n + \sum_{n=1}^{N} \beta'_n x_n^2 + \sum_{n=1}^{N-1} \sum_{n'=n+1}^{N} \beta_{nn'} x_n x_{n'} \quad (15)$$

$$S_{RBF}(\boldsymbol{x}) = \sum_{k=1}^{K} \alpha_k \sqrt{\|\boldsymbol{x} - \boldsymbol{x}^{(k)}\|^2 + \nu^2} + \sum_{n=1}^{N} \beta_n x_n + \beta_0 \quad (16)$$

$$S_{ANN}(\boldsymbol{x}) = a\left[\sum_{n'=1}^{4N} w_{1n'1}\, a\left(\sum_{n=1}^{N} w_{0nn'} x_n + b_{1n'}\right) + b_{21}\right] \quad (17a)$$

$$a(x) = \frac{e^x - e^{-x}}{e^x + e^{-x}} \quad (17b)$$

$$S_{RF}(\boldsymbol{x}) = \frac{1}{T}\sum_{t=1}^{T}\frac{1}{K_t}\sum_{k\in LN_t} y^{(k)} \tag{18}$$

$\alpha$s and $\beta$s in Eq. 15 and 16 are model parameters. $\nu$ in Eq. 16 is a hyperparameter to be fixed before model training. $a(x)$ is the tanh activation function and $w$s and $b$s are the weights and biases in Eq. 17. $K_t$ in Eq. 18 denotes the number of sample points in the leaf node ($LN_t$) of $t^{th}$ decision tree.

For simplicity, we do not specify $MSE^*$. Hence, $\boldsymbol{S}^*$ does not exist. The $dof$s for PRSM and RBF can be derived analytically as $(N+2)(N+1)/2$ and $K$, respectively. However, for ANN and RF, $dof$ must be estimated empirically using the procedure discussed earlier in Section 3. We use Sobol sampling (Sobol, 1967) to generate the train and test sets with 200 sample points each. We take $MSE$ on the data set as the measure of predictive accuracy. To compute $dof$ for ANN and RF, we add Gaussian white noise with $\sigma_\epsilon^2 = 0.1s_y^2$ to the function evaluations at the train points to generate $B = 100$ distinct response sets and estimate $dof$ using Eq. 2. Table 1 lists $s_y^2$, $MSE_m$, and $dof_m$ for each model. Table S1 in the supplementary file presents the pseudo-code to reproduce the data and results.

Let us compute $SQS_m$ for various $\gamma$ to understand the accuracy-complexity trade-offs. For $\gamma = 0$, *SQS* becomes a purely accuracy-based metric, and our procedure gives RBF as the best model with $SQS_{RBF} = 0.090$, followed by RF (0.100), ANN (0.150), and PRSM (0.587). For $\gamma = 0.2$, $SQS$ deemphasizes predictive accuracy slightly, but model ranks remain unchanged from $\gamma = 0$. However, the gap between RBF and RF narrows because RF has much lower complexity than RBF ($dof_{RF} = 122.8$ vs $dof_{RBF} = 200$). On the other extreme of $\gamma = 1.5$, *SQS* prefers model simplicity strongly over predictive accuracy. Here, ANN achieves the least *SQS* (0.078), followed by RF (0.087), RBF (0.090), and PRSM (0.480). Although ANN has higher $MSE$ (0.955) than RBF (0.577) and RF (0.640), it has substantially lower $dof$ (30) than RBF (200) and RF (122.8), which results in the lowest *SQS*. Lower *dof* also makes RF a better choice than the most accurate RBF. Despite being the simplest, PRSM gets the highest $SQS$ due to its poor accuracy. More balanced trade-offs using $\gamma = 0.6$ and $\gamma = 1.1$ yield different outcomes. $\gamma = 0.6$ gives the same model ranks $\gamma = 0$. $\gamma = 1.1$ makes RF the best model by a slim margin over RBF ($SQS_{RF} = 0.0902$ vs $SQS_{RBF} = 0.0904$), followed by ANN (0.093), and PRSM (0.195).

To fully understand the impact of $\gamma$, we compute $SQS$ for each model with $\gamma = 0, 0.1, 0.2, \ldots, 2.5$ and analyze the model ranks. Fig. 6a plots $SQS_m$ vs. $\gamma$ for each model, while Table 2 lists $SQS_m$ at selected $\gamma$s. $SQS$ for RBF remains constant at 0.0904 as it has the highest $dof$, resulting in $CF = 1$. For all other models, $SQS$ decreases with $\gamma$ as model simplicity is preferred more and more. Model ranks are insensitive over a wide range ($0 \leq \gamma \leq 1.07$) of $\gamma$ (see Table 2 and Fig. 6a). RBF, RF, and ANN exhibit very similar $SQS$ values within $1.0 \leq \gamma \leq 1.3$, hence Fig. 6b depicts a zoomed-in $SQS$ plot for this window. For $1.08 \leq \gamma \leq 1.16$, RF outperforms RBF, followed by ANN and PRSM (see Fig. 6b). It continues to outperform RBF for $\gamma \geq 1.08$. For $1.17 \leq \gamma \leq 1.18$, RF remains the best, but ANN outperforms RBF (Fig. 6b) due to its lower $dof$ and it overtakes RF at higher $\gamma$s (see Fig. 6ab). Thus, ANN is preferred for $1.19 \leq \gamma \leq 2.4$. Beyond $\gamma = 2.41$, PRSM overtakes all other models due to its lowest $dof$ (see Table 2 and Fig. 6a). For $\gamma \geq 1.9$, RBF becomes the worst model.

The above example shows that $\gamma$ can impact model selection and users must carefully select it. For this example, the ranks are nearly insensitive unless $\gamma$ is high.

## 7. Conclusion

This work addressed an oft-ignored issue of accuracy versus complexity in the selection of a surrogate model. It presented a novel, composite, hybrid performance metric, Surrogate Quality Score ($SQS$) to integrate these two critical model attributes into a unified quantitative and objective score. $SQS$ combines the two attributes in a simple weighted product of normalized measures and offers a user-specified parameter to adjust the accuracy-complexity trade-off, thereby incorporating human judgement and intuition into the model selection process. It can be used with all forms of surrogate models.

While an illustrative example showed the impact of the user-specified parameter on model selection, results can change significantly in real systems with much uncertainty. Therefore, *SQS* allows the user to play a key role in model selection, as any procedure should. In this sense, SQS offers a simple, practical, and objective approach for model selection by bridging data-driven decision-making with human judgement.

**Acknowledgement**

This study was supported partly by the Agency for Science, Technology and Research, Singapore under its Low-Carbon Energy Research Funding Initiative (U2102d2003) and partly by Pfizer Pte Ltd Singapore under the Pharma Innovation Programme Singapore (PIPS) project (Grant Number: A-0009199-01-00).

**Supplementary Information**

The supplementary information for this work can be found in the online version at DOI: .

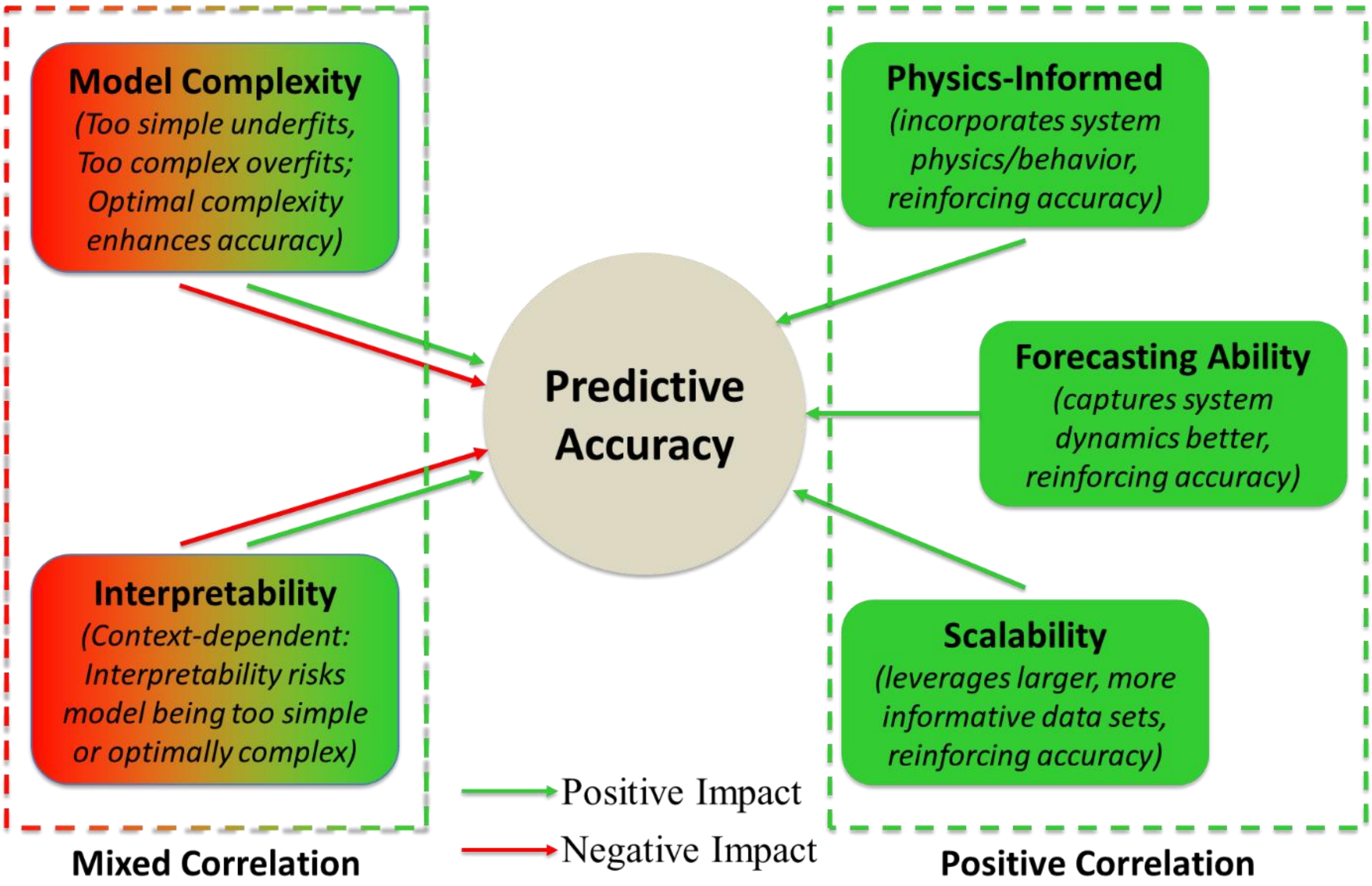

**Fig. 1:** Correlation between predictive accuracy and other desirable traits for model selection

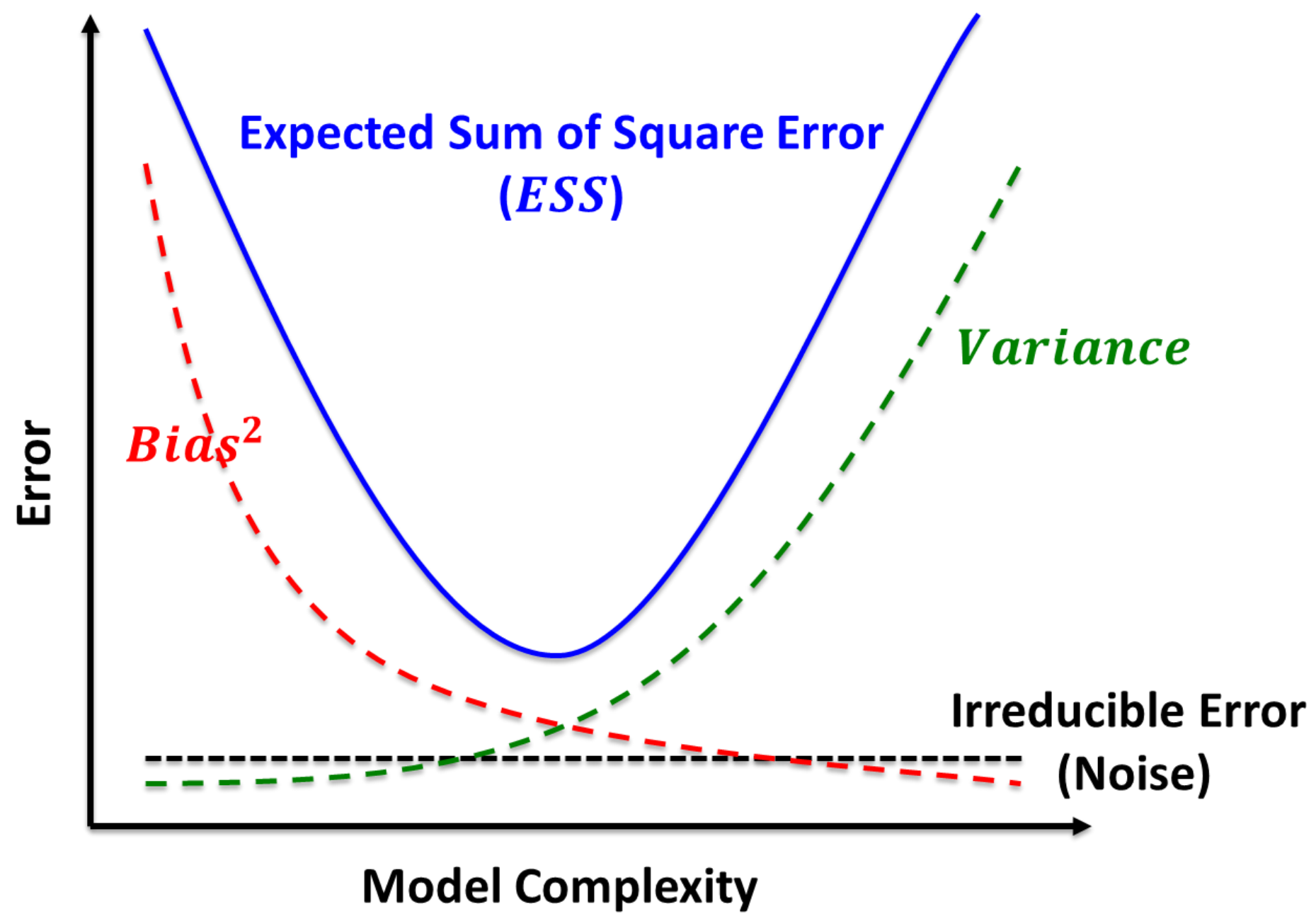

**Fig. 2:** U-shaped Expected Sum of Squared curve along with its individual components: $Bias^2$, $Variance$, and Irreducible Noise

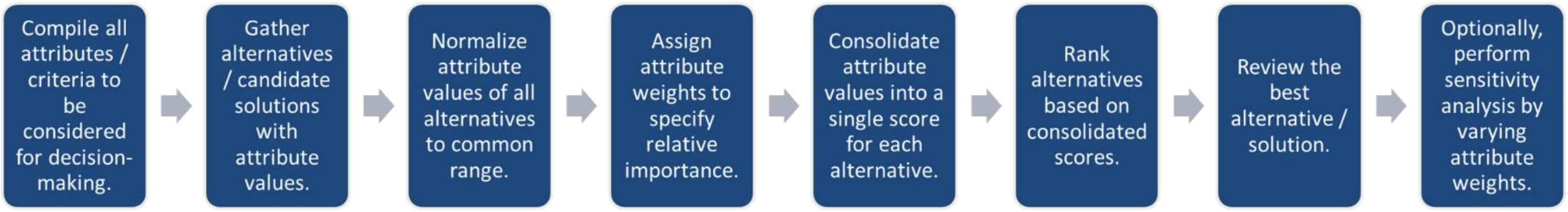

**Fig. 3:** Typical steps in an MCDM/MADM exercise

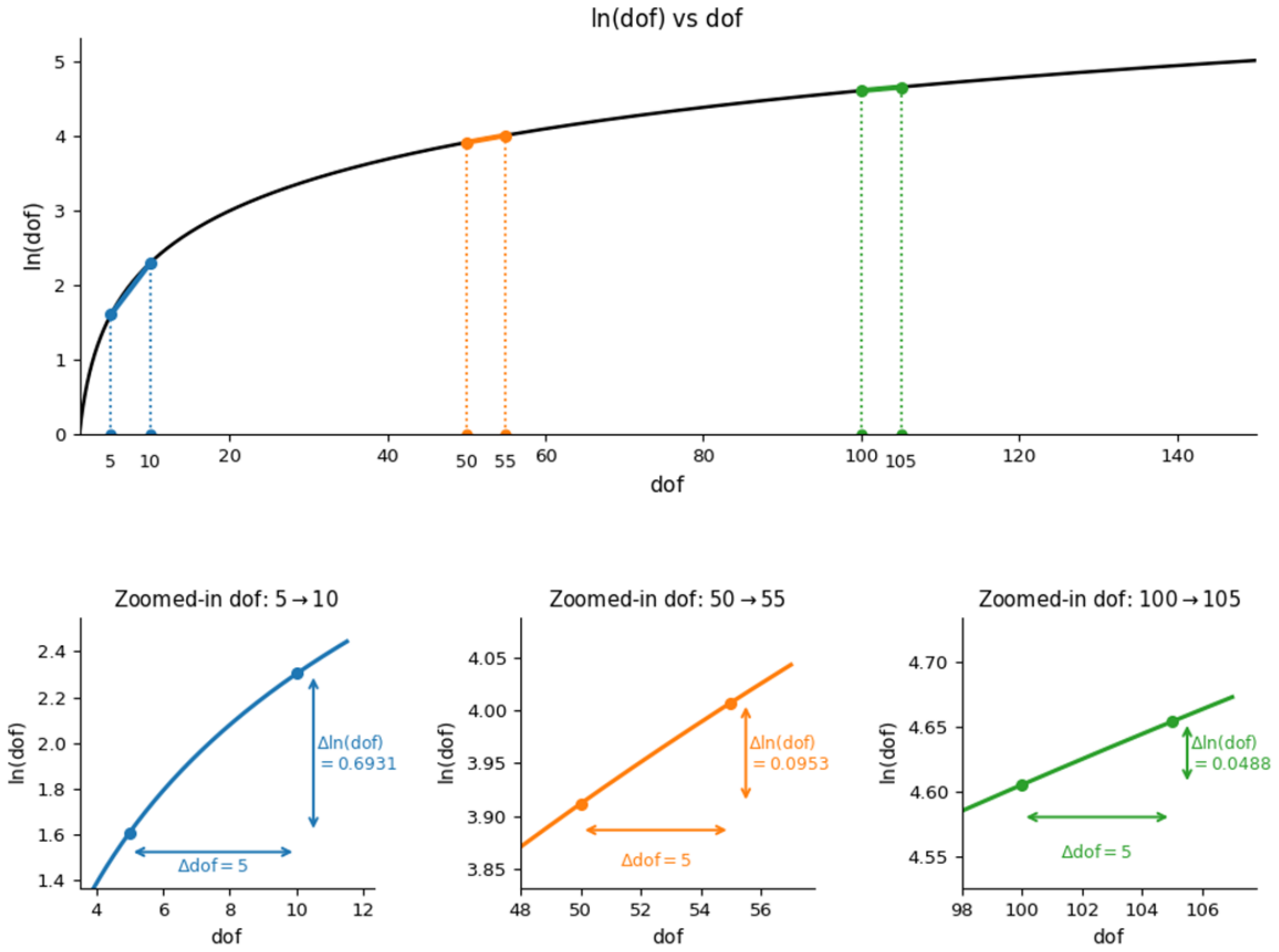

**Fig. 4:** $ln(dof)$ vs $dof$ plot to illustrate the impact of increasing $dof$ on perceived model complexity

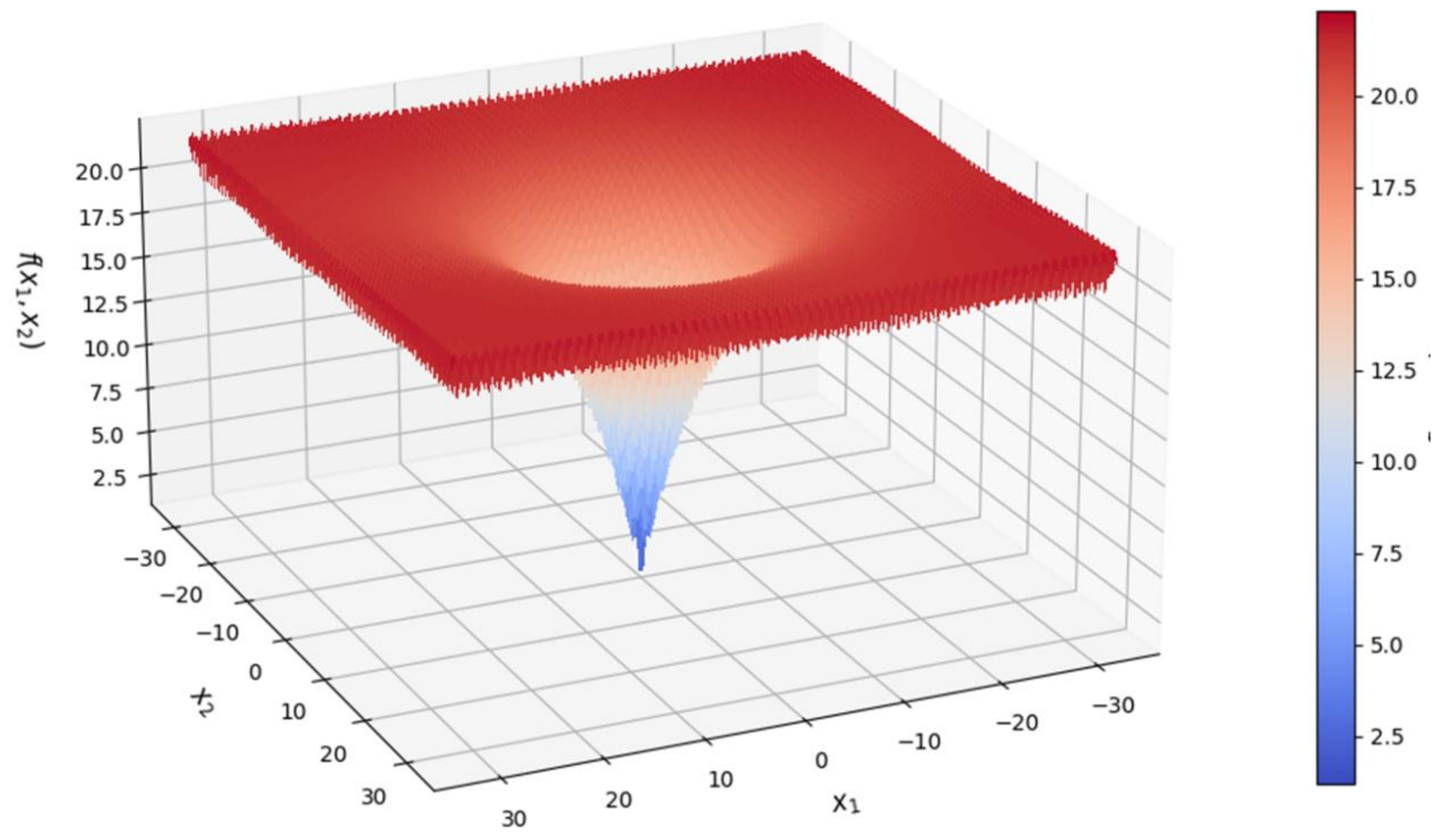

**Fig. 5:** Response Surface for the 2-dimensional Ackley function

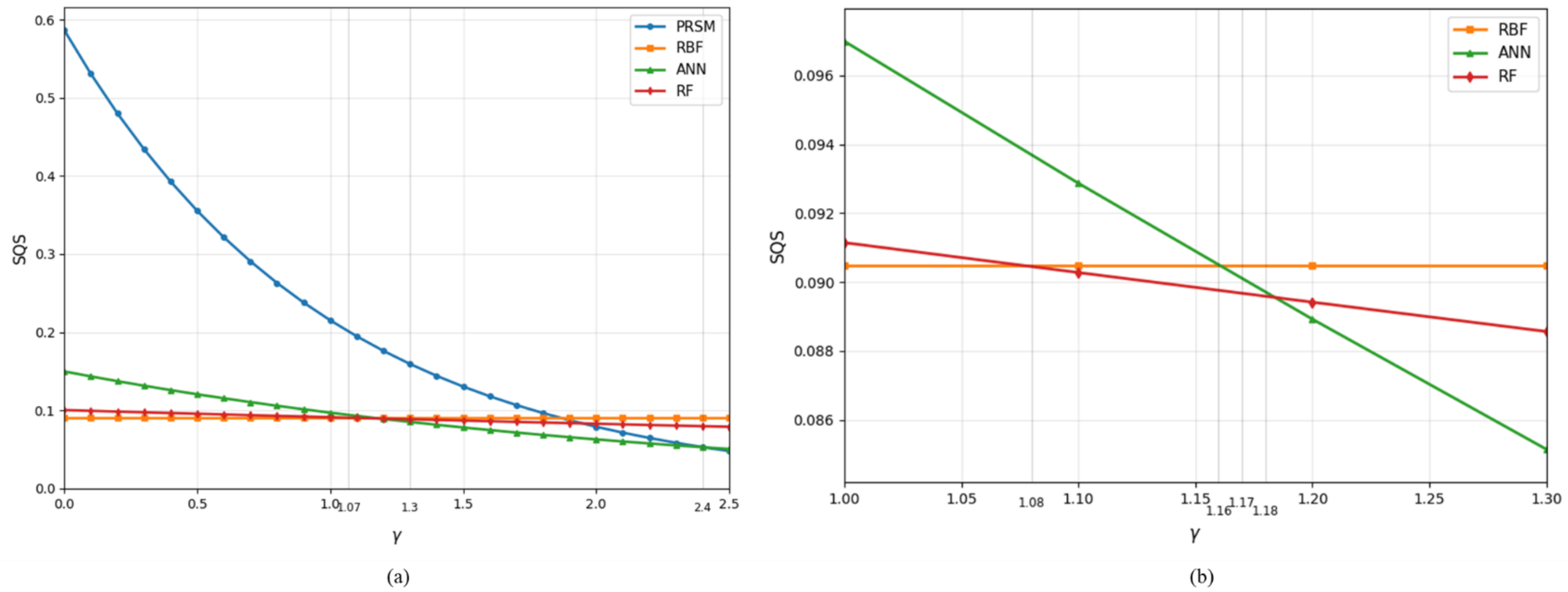

**Fig. 6: (a)** $SQS_m$ vs $\gamma$ for all models over $\gamma \in [0, 2.5]$, **(b)** Zoomed-in $SQS_m$ vs $\gamma$ plot for RBF, ANN, and RF over $\gamma \in [1, 1.3]$

**Table 1:** $MSE$ and $dof$ for each model in the illustrative example

| $s_y^2$ | Model | $MSE_m$ | $dof_m$ |
|---|---|---|---|
| 6.375 | PRSM | 3.739 | 6 |
| | RBF | 0.577 | 200 |
| | ANN | 0.955 | 30 |
| | RF | 0.640 | 122.8 |

**Table 2:** $SQS$ vs $\gamma$ for the four modeling techniques

| Gamma | $SQS$ | | | |
|---|---|---|---|---|
| | **PRSM** | **RBF** | **ANN** | **RF** |
| 0 | 0.5865 | 0.0905 | 0.1498 | 0.1003 |
| 0.1 | 0.5306 | 0.0905 | 0.1434 | 0.0994 |
| 0.2 | 0.4800 | 0.0905 | 0.1373 | 0.0984 |
| 0.3 | 0.4342 | 0.0905 | 0.1315 | 0.0975 |
| 0.4 | 0.3928 | 0.0905 | 0.1259 | 0.0965 |
| 0.5 | 0.3553 | 0.0905 | 0.1205 | 0.0956 |
| 0.6 | 0.3214 | 0.0905 | 0.1154 | 0.0947 |
| 0.7 | 0.2907 | 0.0905 | 0.1105 | 0.0938 |
| 0.8 | 0.2630 | 0.0905 | 0.1058 | 0.0929 |
| 0.9 | 0.2379 | 0.0905 | 0.1013 | 0.0920 |
| 1 | 0.2152 | 0.0905 | 0.0970 | 0.0911 |
| 1.07 | 0.2006 | 0.0905 | 0.0941 | 0.0905 |
| 1.08 | 0.1986 | 0.0905 | 0.0937 | 0.0904 |
| 1.1 | 0.1947 | 0.0905 | 0.0929 | 0.0903 |
| 1.16 | 0.1833 | 0.0905 | 0.0905 | 0.0898 |
| 1.17 | 0.1815 | 0.0905 | 0.0901 | 0.0897 |
| 1.18 | 0.1797 | 0.0905 | 0.0897 | 0.0896 |
| 1.19 | 0.1779 | 0.0905 | 0.0893 | 0.0895 |
| 1.2 | 0.1761 | 0.0905 | 0.0889 | 0.0894 |
| 1.3 | 0.1593 | 0.0905 | 0.0851 | 0.0886 |
| 1.4 | 0.1441 | 0.0905 | 0.0815 | 0.0877 |
| 1.5 | 0.1304 | 0.0905 | 0.0781 | 0.0869 |
| 1.6 | 0.1179 | 0.0905 | 0.0747 | 0.0861 |
| 1.7 | 0.1067 | 0.0905 | 0.0716 | 0.0852 |
| 1.8 | 0.0965 | 0.0905 | 0.0685 | 0.0844 |
| 1.9 | 0.0873 | 0.0905 | 0.0656 | 0.0836 |
| 2 | 0.0790 | 0.0905 | 0.0628 | 0.0828 |
| 2.1 | 0.0714 | 0.0905 | 0.0601 | 0.0820 |
| 2.2 | 0.0646 | 0.0905 | 0.0576 | 0.0812 |
| 2.3 | 0.0585 | 0.0905 | 0.0551 | 0.0805 |
| 2.4 | 0.0529 | 0.0905 | 0.0528 | 0.0797 |
| 2.41 | 0.0523 | 0.0905 | 0.0526 | 0.0796 |
| 2.5 | 0.0478 | 0.0905 | 0.0505 | 0.0789 |